\pdfoutput=1

\documentclass[11pt]{article}

\usepackage{acl}

\usepackage{times}
\usepackage{latexsym}
\usepackage{graphicx}
\usepackage{amsmath}
\usepackage{hyperref}
\usepackage{booktabs} 
\usepackage{multirow} 
\usepackage{multicol} 
\usepackage{subfig}
\usepackage{CJKutf8}
\usepackage{algorithm}
\usepackage{algorithmic}
\usepackage{tabularx}

\usepackage[T1]{fontenc}

\usepackage[utf8]{inputenc}

\usepackage{microtype}

\usepackage{inconsolata}

\title{Think from Words(TFW): Initiating Human-Like Cognition in Large Language Models Through Think from Words for Japanese Text-level Classification}

\author{
  Chengguang Gan\textsuperscript{1*} \quad
  Qinghao Zhang\textsuperscript{2*} \quad
  Tatsunori Mori\textsuperscript{1} \\
  \textsuperscript{1}Yokohama National University, Japan \\
  \texttt{gan-chengguan-pw@ynu.jp, tmori@ynu.ac.jp} \\
  \textsuperscript{2}Department of Information Convergence Engineering, \\
  Pusan National University, South Korea \\
  \texttt{zhangqinghao@pusan.ac.kr} 
  \thanks{\textsuperscript{*}Equally Contribution.}
}

\begin{document}
\maketitle

\begin{abstract}

The proliferation of Large Language Models (LLMs) has spurred extensive research into LLM-related Prompt investigations, such as Instruction Learning (IL), In-context Learning (ICL), and Chain-of-Thought (CoT). These approaches aim to improve LLMs' responses by enabling them to provide concise statements or examples for deeper contemplation when addressing questions. However, independent thinking by LLMs can introduce variability in their thought processes, leading to potential inaccuracies. In response, our study seeks to bridge the gap between LLM and human-like thinking processes, recognizing that text comprehension begins with understanding individual words. To tackle this challenge, we have expanded the CoT method to cater to a specific domain. Our approach, known as "Think from Words" (TFW), initiates the comprehension process at the word level and then extends it to encompass the entire text. We also propose "TFW with Extra word-level information" (TFW Extra), augmenting comprehension with additional word-level data. To assess our methods, we employ text classification on six Japanese datasets comprising text-level and word-level elements. Our findings not only validate the effectiveness of TFW but also shed light on the impact of various word-level information types on LLMs' text comprehension, offering insights into their potential to cause misinterpretations and errors in the overall comprehension of the final text.

\end{abstract}

\section{Introduction}

Since Large Language Models (LLMs) inception \citep{zhao2023survey}, LLMs have garnered substantial attention across various academic domains. Indeed, LLMs have rapidly emerged as the prevailing focal point within the realm of natural language processing research. Commencing with the pioneering GPT-3 \citep{brown2020language}, followed by innovations like BLOOM \citep{workshop2022bloom}, ChatGPT \citep{ouyang2022training}, and the groundbreaking GPT-4 \citep{openai2023gpt4}, these models have unequivocally ushered in the era of LLMs. Furthermore, open-source LLM LLaMA \citep{touvron2023llama} \citep{touvron2023llama2} has gained widespread recognition as a foundational model, serving as the bedrock for numerous applications. Central to the potency of LLMs is their utilization of extensive corpora for pre-training, encompassing billions to hundreds of billions of parameters. This pre-training strategy bestows upon LLMs remarkable generalization and reasoning capabilities, thereby augmenting their overall effectiveness.

Conversely, the challenge of optimizing the use of LLMs has gained prominence. Since its inception, the concept of "Prompt" has been integral in facilitating more efficient interactions with LLMs. In-context learning (ICL) \citep{brown2020language}, initially introduced for GPT-3 models, represents a pioneering Prompt methodology. This approach involves presenting a question-answer pair, followed by a similar question. The LLM then processes these three texts collectively, learning to generalize the response mechanism based on the provided examples. Consequently, the LLM produces an answer to the second question, echoing the format of the initial pair. The advent of ICL and the evolution of GPT-3 underscore a significant milestone: post-expansion in parameter size, LLMs exhibit learning capabilities reminiscent of human cognition. They can address a series of similar questions by referencing a single exemplar. Subsequently, Instruction Learning (IL) emerged \citep{wei2021finetuned}, directing models to generate specific responses based on explicit instructions. The introduction of IL signifies the LLMs' comprehension of human directives and their ability to respond accordingly.

The validation of LLMs' learning and instructional adherence through ICL and IL was further advanced with the introduction of Chain-of-Thought (CoT) reasoning \citep{wei2023chainofthought}. CoT, a problem-solving and decision-making method, significantly augments the reasoning capabilities of LLMs. It involves a step-by-step breakdown of complex problems into smaller, more manageable segments. This process not only simplifies problem-solving but also fosters coherent solutions and understandings, ultimately guiding the model to the correct answer. The CoT method has catalyzed a surge in related research. Nevertheless, these studies have not yet delved into the extent to which LLMs can emulate human-like understanding. To truly grasp the overall meaning of a text, one must begin with a comprehension of each word, a principle that mirrors human learning processes.

\begin{figure}[!t]
\centering
\includegraphics[width=219 pt]{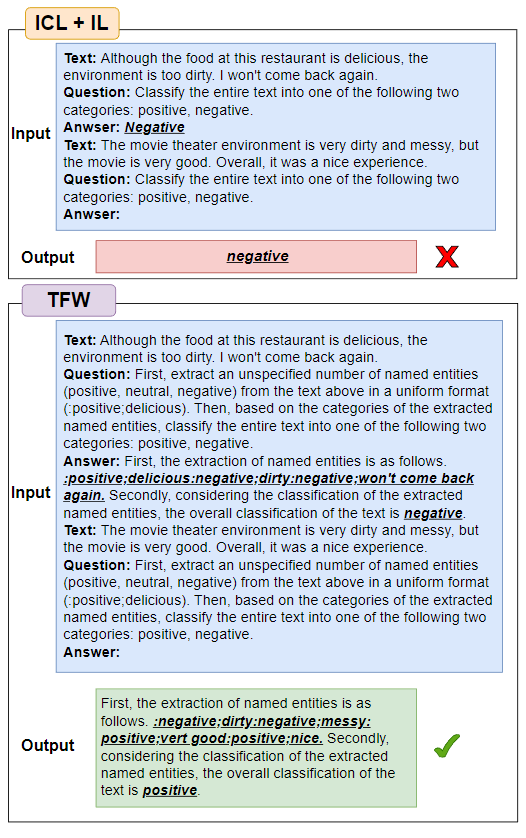}
\caption{\label{1figure1}The illustration depicts the In-context Learning + Instruction Learning method and Think from Words method in same sentiment classification tasks.}

\end{figure}

In summary, this study focuses on a specific problem within the realm of natural language understanding, which is the text-level classification task. This task metaphorically evaluates whether a model can grasp the overall meaning of a text. It is intricately linked with word-level classification, where the comprehension of individual words is paramount. The central inquiry is whether enhancing a model's word-level understanding can subsequently improve its comprehension of the entire text. This exploration aims to determine if LLMs can emulate human-like text understanding, beginning with the interpretation of each word in the text.

The manuscript presents a binary classification task for sentiment analysis, as depicted in Figure \ref{1figure1}, utilizing the ICL plus IL approach. Initially, the process involves presenting a text segment, followed by a guiding question, and culminating with a response to this question. The first three text examples serve as instructional cases for ICL learning in Large Language Models (LLMs). Subsequently, the pertinent text and question for classification are combined. The final step involves procuring the model's output. However, challenges arise, notably with the provided examples being negatively biased. For instance, although the initial segment of a text may convey negative sentiment, its latter part might be positive, leading to an overall positive sentiment. Despite this, the model erroneously categorizes such texts as negative. This misclassification stems from the model's superficial word-level analysis within sentences. 

To address this issue, this study introduces the Think from Words (TFW) method, which significantly mitigates this problem by enabling the model to initially comprehend word-level semantics and subsequently infer text-level classification based on this understanding. This approach aligns with the comprehensive understanding of the entire text. It is important to clarify that the CoT method is not included in this study, as TFW is posited as an extension of CoT, specifically tailored for text classification tasks. The fundamental proposition of this paper is to encourage the model to evaluate the text holistically, akin to human reading practices, before delving into individual word analysis. This concept forms the core of the TFW methodology and embodies the primary objective of this research. This paper presents several key contributions:

1. Introduction of the Think from Words (TFW) methodology. This approach illustrates how LLMs can process text akin to human reading, beginning with word-level comprehension and progressing to a more sophisticated grasp of overall textual meaning.

2. Development of the TFW with Extra word-level information (TFW Extra) method. This enhanced method further refines LLMs' capability to understand entire texts by integrating supplementary external information, thereby expanding their interpretative scope.

3. Comprehensive analysis involving diverse datasets and external word-level information types. This study elucidates how certain types of word-level data can lead LLMs to misclassify textual content, while others may aid in improving text-level classification accuracy. The findings also reveal that LLMs, when supplemented with additional word-level information, exhibit behavioral patterns consistent with fundamental human cognitive processes in their responses to this data.

\section{Related Work}

The previous chapter provided a comprehensive analysis of three pivotal studies: ICL, IL, and CoT, which are crucial for this research. However, recent developments in related fields warrant attention. This section aims to present works pertinent to our study, categorizing them into two distinct groups. The first category encompasses research that leverages the intrinsic knowledge of LLMs as the primary driver of the investigation. The second category includes studies that incorporate external knowledge sources to augment the capabilities and performance of LLMs.

\textbf{Intrinsic Knowledge.} \citet{zhang2022automatic} presents Auto-CoT, an automated method for enhancing large language models' reasoning by generating diverse, self-constructed demonstrations, outperforming manual approaches in reasoning tasks. \citet{wang-etal-2022-iteratively} introduces an iterative prompting framework to enhance PLMs for complex, multi-step reasoning tasks. \citet{wu-etal-2023-self} developed a self-adaptive framework for in-context learning in language models, significantly improving example selection and ordering. Recent studies have advanced the CoT approach, yielding significant enhancements in the reasoning capabilities of LLMs \citep{huang2022large} \citep{zhou2023leasttomost} \citep{wang2022rationale} \citep{wang2023selfconsistency}. Reliance on language models' self-generated knowledge often leads to inaccuracies, as reasoning based on erroneous information inevitably produces incorrect conclusions, a phenomenon known as illusions in language models \citep{huang2023survey} \citep{ji2023survey}.

\textbf{External Knowledge.} To address the hallucination issue inherent in Large Language Models (LLMs), it is imperative to integrate more dependable external knowledge sources. A practical approach to achieving this is through the incorporation of knowledge graphs. By amalgamating LLMs with these knowledge graphs \citep{agrawal2023can}, which encompass accurate and verified information, we can effectively mitigate the generation of erroneous knowledge \citep{agarwal2023bring} \citep{wang2023knowledge} \citep{wang2023boosting} \citep{baek2023knowledge} \citep{trivedi2022interleaving}. This integration not only enhances the reliability of LLMs but also enriches their knowledge base, leading to more accurate and trustworthy outputs. After there, \citet{tian2023graph} introduces Graph Neural Prompting (GNP), a method enhancing Large Language Models' knowledge grounding using knowledge graphs without retraining, proving effective in various reasoning tasks. 

Contrasting with prior research, this paper centers on the exploration of LLMs' proficiency in comprehending both word-level and text-level nuances. This investigation is grounded in the hypothesis that to emulate human-like cognitive processing in reading and interpreting texts, LLMs must initially acquire a nuanced understanding of individual words. Such foundational comprehension is crucial for these models to effectively grasp and interpret the broader context and meaning of complete texts.

\section{Think from Words}

The upcoming section of this paper is structured to provide a comprehensive understanding of the components constituting "Think from Word" (TFW). Initially, we will delineate the individual elements that form the foundation of TFW. This will be followed by an in-depth examination of the design and function of each segment within the text structure of TFW. Concluding this section, we will discuss methodologies for incorporating external word-level information into the basic TFW framework, aimed at enhancing the text-level understanding capabilities of LLMs.

\begin{figure*}[!t]
\centering
\includegraphics[width=440 pt]{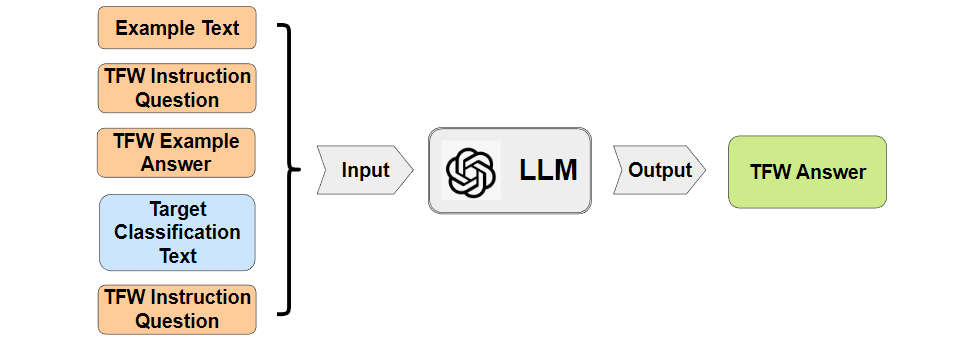}
\caption{\label{2figure2}The illustration depicts the components of input and output in Think from Words (TFW).}

\end{figure*}

\begin{figure*}[!t]
\centering
\includegraphics[width=440 pt]{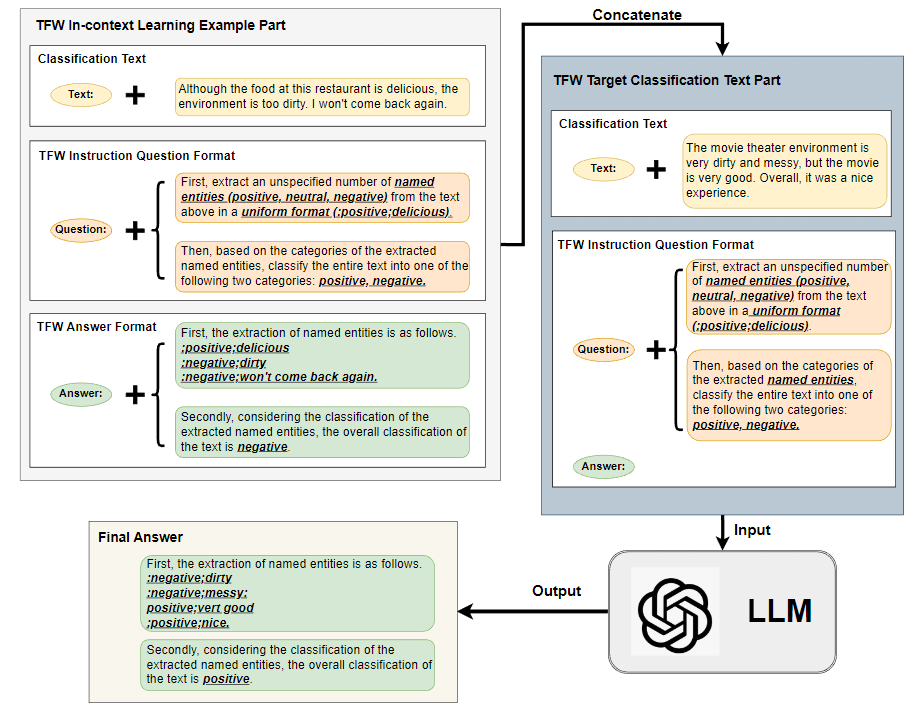}
\caption{\label{3figure3}The illustration depicts the detailed design of the text of each part of the TFW in the sentiment classification task.}

\end{figure*}

The process of TFW is illustrated in Figure \ref{2figure2}, where each text component is integrated into a LLM. The crux of this process is the model's input, which is composed of five distinct parts. Initially, the text example is introduced, followed by the TFW formatted instruction question and a corresponding example answer. These three elements collectively form the instructional sample for the model's learning process, analogous to ICL. Subsequently, the target text requiring categorization is added. This is followed by the repetition of the TFW instruction question, mirroring the earlier example. These five modules are sequentially combined and input into the LLM to generate a answer with TFW format. Notably, each input sample consistently contains four fixed elements, depicted as orange blocks, while the blue blocks represent varying target texts for classification. The LLM outputs the response in a format akin to the provided example, represented by the green block in Figure \ref{2figure2}. While the input format superficially resembles a combination of ICL and IL, the essence of this approach lies in the meticulous design of the TFW instruction questions and answers. The subsequent section will elaborate on the design aspects of the various text components constituting the TFW methodology.

\subsection{Design of TFW}

This study aims to develop LLMs that emulate human-like cognition, think beginning at the word-level. The process starts by analyzing individual words within a text, building towards a comprehensive understanding of the entire text. To illustrate this concept, consider its application in a natural language processing task. For instance, the model first performs tasks such as extracting and classifying key words in a text, akin to a Named Entity Recognition (NER) task. Subsequently, the model is tasked with categorizing the entire text based on these extracted word-label pairs, similar to a Text Sentiment Classification task. The detailed flow of a basic TFW on text-level classification task is shown in Figure \ref{3figure3}. Here we still use the text sentiment classification task as an example for illustration. The text sentiment classification task tests the model's ability to control the overall understanding of the text.

The \textbf{TFW In-context Learning Example}, located in the upper left corner of the Figure \ref{3figure3}, is structured into three distinct components. The first component presents the text of the ICL example, followed by a \textbf{TFW instructional question}, which is subdivided into two stages for clarity and effectiveness. The initial stage involves instructing the model to identify and extract a varying number of label-word pairs from the given text, utilizing a standardized format. This process is facilitated by the inclusion of a parenthetical note following the term "named entities". Within these parentheses, a comprehensive list of potential categories for word classification is provided. For instance, in tasks involving text sentiment analysis, the categories for word classification are typically delineated as "positive" and "negative." Furthermore, an illustrative label-word pair is appended after "uniform format" to serve as a clear example (e.g. :positive;delicious). It's important to note that the symbols ":," and ";" denote the commencement and conclusion of a label, respectively. This standardized format is adapted from the Japanese word-level and text-level mix classification dataset, a point that will be elaborated upon in the dataset section of the experimental setup.

The second stage of the instruction prompts the model to synthesize the earlier extracted label-word pairs to determine an overall classification for the text, such as "positive" or "negative." Finally, there is the \textbf{TFW answer example}, which follows the format of the previous question to answer the question in the order of the two steps. In summary, these three components collectively establish a comprehensive \textbf{Text-Question-Answer} cycle, constituting a complete ICL example.

The right side of Figure \ref{3figure3} illustrates the \textbf{TFW Target Classification Text Part}. This section involves substituting the existing text with the designated target classification text. Notably, the concluding question aligns precisely with the example previously presented in the ICL section. Sequentially, this approach integrates five distinct text blocks, subsequently processed by LLMs. As a result, the LLMs produce TFW format responses characterized by word-level label-span pairs. The ultimate determination of text classification is achieved by synthesizing these label-span pairs to assess the text at a text-level.

The above description outlines the comprehensive workflow of the TFW model. The initial step in this process involves the model generating classifications for words that are pertinent to the overarching classification of the text. Subsequently, these classifications are leveraged to form strongly correlated label-span pairs. Utilizing these pairs, a more accurate determination can be made regarding the overall classification of the text. The essence of this approach lies in beginning the analysis at the word level and progressively broadening the scope to encompass the entire text. It is important to acknowledge that the types of labels in the label-span pairs may vary depending on the specific word-level classification task at hand, which is often influenced by the characteristics of different datasets.

\begin{figure}[h]
\centering
\includegraphics[width=219 pt]{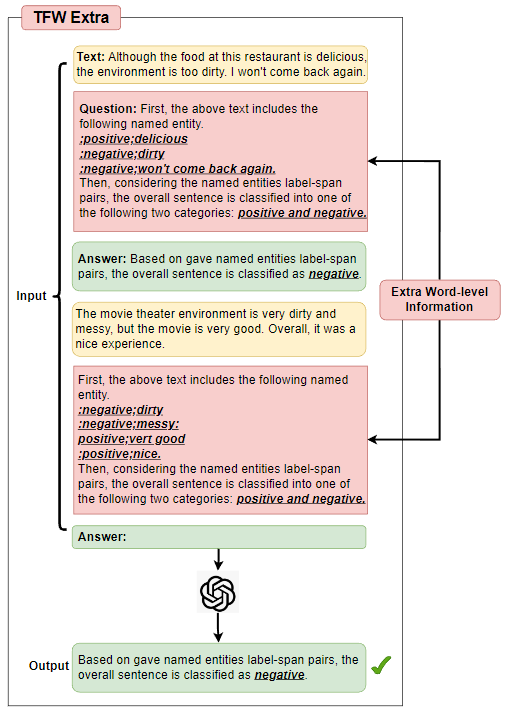}
\caption{\label{4figure4}The illustration depicts the detailed of the TFW Extra word-level information (TFW Extra) method.}

\end{figure}

\subsection{TFW with Extra Word-level Classification Information}

In the preceding section, we discussed the significant issue of hallucinations in LLMs. This raises a crucial query: do inaccuracies in generating label-span pairs by LLMs lead to misclassifications in the overall text analysis? To investigate this, we introduce an advanced approach called TFW with Extra Word-Level Information (TFW Extra). This method involves the integration of manually annotated extra word-level information (e.g. label-span pairs) into the input text. Our aim is to explore how different types of added word-level information enhance or impair the model's comprehension and classification accuracy across various tasks.

Figure \ref{4figure4} illustrates a comparison between TFW and TFW Extra, highlighting that the latter incorporates additional label-span pairs, denoted in red, into the input text sequence. The process begins with modifying the question text to include specific label-span pairs. Following this, the command model categorizes the text based on these predefined pairs. Unlike traditional LLMs that generate label-span pairs independently, this approach utilizes external knowledge to aid the LLMs in responding more accurately. This methodology exemplifies an ICL process, which is designed to facilitate learning in LLMs. The cycle involves selecting relevant classification text and questions, and feeding these elements, along with the label-span pairs, into the model. The final output (e.g., classifying a sentence as negative based on specific named entity label-span pairs) demonstrates the effectiveness of this enhanced approach.

\section{Experiment Setup}

In this section, we provide a comprehensive elucidation of the experimental setup. This encompasses a meticulous description of the criteria for datasets selection, the rationale behind choosing specific models, and the methodologies employed for evaluation.

Given the limitations of time and resources, it was not feasible to test every sample in the dataset. Consequently, we employed a strategy of random selection, choosing 1,000 samples for testing (SCPOS: RW dataset and TCREE dataset are all used as test sets.). To enhance the robustness of our findings, each selected sample was subjected to three experimental cycles. The results of these cycles were then averaged to derive a final value for each sample. To preserve the integrity of randomization while ensuring consistency across the three experimental runs, we implemented a controlled random seed approach. Specifically, we used three distinct random seeds: 42, 123123, and 678910. This methodological choice was aimed at bolstering the reliability and fairness of the experimental outcomes. All experiments used 1-shot as ICL example\footnote{The TCREE datasets are comprised of two distinct types of tasks: Relationship Extraction (RE) and Event Extraction (EE). Consequently, the application of the TCREE datasets involves a 2-shot approach, utilizing one sample each from the RE and EE tasks for comprehensive analysis.}.

\begin{table*}[!t]
\centering
\begin{tabular}{lll}
\hline
 Datasets & \textbf{Text-level} & \textbf{Word-level}  \\
\hline
 SCNM & Society, Literature, Academia, & people, corporations, political\\
  &  Technology, Nature &   organizations, other organizations,\\
&  &   places, facilities, products, and events   \\
\hline
 SCPOS:RW& positive, negative & positive, neutral, negative \\
 \hline
 SCPOS:N& positive, negative & positive, neutral, negative  \\
 \hline
 SCPOS:Adj& positive, negative & positive, negative  \\
 \hline
 SCPOS:N \& Adj& positive, negative & positive, neutral, negative  \\
 \hline
 TCREE & sports, film, women, IT, advertising & affiliation, occupation, starring, director,   \\
  &   &  age, product, goods, performances, wins,  \\
  &   &   broadcasts, public appearances, launches,\\
  &   &   retirements\\
\hline
\end{tabular}
\caption{\label{table1mixdataset}
The table provides label words, delineating the all of label in six mix information extraction Japanese datasets. SCNM: Sentence Classification and Named Entity Recognition Mix Dataset. SCPOS: Sentiment Classification and Part-of-Speech Dataset. RW: Relation Word. N: Noun. Adj: Adjective. N \& Adj: Nous and Adjective. TCREE: Text Classification and Relation \& Event Extraction Dataset.}
\end{table*}

\subsection{Dataset}

In selecting appropriate datasets, the TFW method requires a dataset that encompasses both text-level and word-level label pairs. Presently, the only dataset meeting this criterion is the "Japanese Information Extraction Mix" dataset, previously utilized by \citet{gan2023giellm} in the development of the Japanese Information Extraction Large Language Model (GIELLM). This specific dataset provides comprehensive labeling at both word and text levels for each entry. Table \ref{table1mixdataset} comprehensively enumerates the labels for each of the six distinct datasets, each corresponding to a different task. These datasets are uniquely designed based on the concept of Mutual Reinforcement Effect, establishing a symbiotic relationship between text-level and word-level labels. This alignment makes them particularly suitable for validating the TFW and TFW Extra methodologies proposed in this study.

\subsection{Model}

Upon selecting an appropriate dataset, we faced a significant challenge in choosing a suitable model. Specifically, we found that no existing open-source multilingual LLMs were capable of effectively handling the complexities of the Japanese language. This included various models that had undergone fine-tuning on Japanese corpora, such as the LLaMA2 model \citep{touvron2023llama2}, as well as the natively developed Japanese LLM-jp-13B\footnote{\raggedright\url{https://huggingface.co/llm-jp/llm-jp-13b-v1.0}}. Our evaluations of these Japanese-specific LLMs revealed a consistent inability to generate coherent text, resulting in outputs that were predominantly irrelevant or nonsensical. Consequently, after thorough consideration, OpenAI's GPT-3.5-Turbo\footnote{\raggedright\url{https://platform.openai.com/docs/guides/text-generation/chat-completions-api}} emerged as the only viable option for testing our TFW method.

\subsection{Evaluation}

Generative models, in contrast to sequence labeling models, produce text sequences that are inherently indeterminate. This is particularly evident in LLMs, where few-shot scenarios are common. As a result, even when explicit instructions and formats are provided through ICL, LLMs do not consistently generate text sequences as text label and label-span pairs. To address this issue, we have developed a straightforward algorithm, outlined in Algorithm \ref{alg:evaluate_accuracy}. This algorithm initially employs regularization to extract the corresponding text label and label-span pairs from the generated responses. Subsequently, these generated labels are compared individually with the actual labels to determine the final accuracy. It is crucial to distinguish between three types of accuracy: Text Classification Accuracy (TC), which measures the overall accuracy of text classification; Label-Span Accuracy (LS), which assesses the accuracy of label-span pairs; and Total Accuracy, calculated when both TC and LS accuracies reach a value of 1. This differentiation allows for a nuanced comparison and analysis of the TFW method's efficacy across various datasets.

\begin{table*}[!t]
\centering

\begin{tabularx}{1\textwidth}{@{}l*{9}{>{\centering\arraybackslash}X}@{}}
\toprule[2pt]
 & \multicolumn{3}{c}{SCNM} & \multicolumn{3}{c}{SCPOS: RW} & \multicolumn{3}{c}{SCPOS: Adj \& N} \\
 Accuracy &  TC &  LS &  Total  & TC &  LS &  Total L& TC &  LS &  Total \\
\midrule
 ICL+IL& 43.73 &  -  &  -  & 73.90 &  -  &  -  & 81.13 &  -  & -  \\
 TFW & 45.47 & 41.45 & 10.83 & \textbf{81.95} & 24.39 & 5.45 & \textbf{85.90} & 21.99 & 14.63 \\
 TFW Extra &  \textbf{46.37} &  -  &  -  & 73.95 &  -  &  -  & 56.6 &  -  & -  \\
\bottomrule[2pt]
\end{tabularx}
\vspace{1em}

\begin{tabularx}{1\textwidth}{@{}l*{9}{>{\centering\arraybackslash}X}@{}}
\toprule[2pt]
\toprule[2pt]
 & \multicolumn{3}{c}{SCPOS: N} & \multicolumn{3}{c}{SCPOS: Adj} & \multicolumn{3}{c}{TCREE} \\
 Accuracy &  TC &  LS &  Total  & TC &  LS &  Total& TC &  LS &  Total \\
\midrule
 ICL+IL& \textbf{81.13} &  -  &  -  & 81.13 &  -  &  -  & \textbf{84.24} &  -  & -  \\
 TFW & 80.30 & 8.92 & 0.30 & \textbf{86.40} & 21.91 & 14.43 & 80.57 & 6.93 & 6.33 \\
 TFW Extra & 46.63 &  -  &  -  & 56.27 &  -  &  -  & 83.89 &  -  & -  \\

\bottomrule[2pt]
\end{tabularx}
\vspace{1em}

\caption{\label{table2results}
The accuracy of Six Mix Datasets in different Method.}
\end{table*}

\section{Results}

The data presented in Table \ref{table2results} elucidates the performance accuracy of six datasets when applied to the GPT-3.5-Turbo model. Initially, the focus is on the Sentence Classification and Named Entity Recognition Mix (SCNM) dataset. Utilizing only the ICL and IL methods, the TL accuracy is recorded at 43.73. However, this accuracy experiences an increase to 45.47 upon the integration of the TFW method. Further enhancement is observed with the addition of extra word-level information, elevating the TL accuracy to 46.37. These incremental improvements are clearly delineated in the results. In the realm of general knowledge text categorization, prompting LLMs to initially consider entities such as names of people and organizations within the text proves beneficial. This strategy aids LLMs in more effectively making overarching classification judgments about the text.

In this study, we evaluated four Sentiment Classification and Part-of-Speech (SCPOS) datasets using the TFW method. Our findings reveal that the TFW method outperforms the baseline ICL and IL methods in terms of TL accuracy across three of the four sub-datasets, with the exception of the SCPOS: N sub-dataset. Notably, in the sub-dataset focusing on sentiment-related words, there was a significant increase in TL accuracy, rising from 73.90 to 81.95. Similarly, the sub-dataset concentrating on sentiment nouns exhibited an improvement in TL accuracy from 81.13 to 85.90. The most substantial enhancement was observed in the sub-dataset dedicated to sentiment adjectives, where TL accuracy improved by 5.27.

These results suggest a notable trend in sentiment analysis of texts: adjectives play a more pivotal role in determining the overall sentiment compared to related words and nouns. This observation aligns with human cognition, indicating that the sentiment polarity of a text is heavily influenced by the sentiment polarity of individual adjectives within it. This insight is crucial for refining approaches in sentiment analysis, emphasizing the importance of weighting adjectives more heavily in such models.

Conversely, applying the TFW method to the SCPOS: N dataset results in a marginal reduction in TL accuracy, notably manifesting as extremely low LS accuracy. Moreover, incorporating additional word-level information across all four SCPOS sub-datasets precipitates a marked decline in TL accuracy. We propose that this phenomenon may be attributed to the inherent sentiment polarity of nouns. Predominantly, nouns exhibit a neutral sentiment (According to the labels in Table \ref{table1mixdataset}, all three sub-datasets of SCPOS except for adjectives have neutral classifications). Consequently, this predisposes language models (LLMs) to interpret the overall sentiment of the text as neutral, culminating in a uniform classification of the text as neutral. However, such an approach conflicts with text categorization systems, which typically do not recognize 'neutral' as a category. This discrepancy significantly impairs TL accuracy.

In the analysis of the Text Classification and Relation \& Event Extraction (TCREE) Dataset, it was observed that the TFW method did not demonstrate superiority over the baseline method. Notably, the LS accuracy for the TFW approach was the lowest among the six datasets examined. This contributed to a reduction in the overall TL accuracy. However, the introduction of additional word-level information in the TFW Extra method led to a significant improvement, with TL accuracy increasing to 83.89. This enhancement underscores the hypothesis that LLMs process information in a word-centric manner, akin to human cognitive processes, thereby facilitating a deeper comprehension of the text in its entirety. Consequently, the integration of supplementary information can be instrumental in augmenting the overall understanding of LLMs.

\section{Conclusion and Future Work}

In this study, we investigate the capability of LLMs to emulate human-like thinking. Recognizing that comprehensive text understanding begins at the word-level, we introduce two novel methodologies: TFW and TFW Extra approaches. We assess the efficacy of the TFW method through a series of experimental evaluations across diverse datasets. Our findings not only validate the effectiveness of the TFW method but also substantiate the premise that LLMs can, akin to humans, initiate thought processes beginning at the word level.

\section{Limitations}

This study encounters limitations primarily in dataset selection, owing to the requirement for specialized text-level and word-level mixed datasets. Consequently, the analysis was confined to six Japanese datasets. Additionally, budget constraints precluded the use of the GPT-4 API. Therefore, the evaluation was limited to the GPT-3.5-Turbo model.

\bibliography{custom}

\appendix
\section{Appendix}

\subsection{Calculate Accuracy Algorithm}\label{algorithm code}

\begin{algorithm}
\caption{Calculate Accuracy Algorithm}
\label{alg:evaluate_accuracy} 
\begin{algorithmic}
\REQUIRE $generated$, $actual\_tc$, $actual\_ls$
\ENSURE $tc\_acc$, $ls\_acc$, $total\_acc$

\STATE $(gen\_tc, gen\_ls) \leftarrow extract\_text\_piece(generated)$
\STATE $tc\_acc \leftarrow 0$
\IF{$gen\_tc$ is not None}
    \STATE $tc\_acc \leftarrow \text{int}(gen\_tc == actual\_tc)$
\ENDIF

\STATE $ls\_acc \leftarrow 0$
\IF{$gen\_ls$ is not None}
    \STATE $gen\_ls\_pairs \leftarrow parse\_label\_ls\_pairs(gen\_ls)$
    \STATE $actual\_ls\_pairs \leftarrow parse\_label\_ls\_pairs(actual\_ls)$
    \STATE $matching\_pairs \leftarrow \text{length of }(gen\_ls\_pairs \cap actual\_ls\_pairs)$
    \IF{$actual\_ls\_pairs$ is not empty}
        \STATE $ls\_acc \leftarrow \text{matching\_pairs} / \text{length of }actual\_ls\_pairs$
    \ENDIF
\ENDIF

\STATE $total\_acc \leftarrow \text{int}(sc\_acc == 1 \text{ and } ls\_acc == 1)$
\RETURN $(tc\_acc, ls\_acc, total\_acc)$
\end{algorithmic}
\end{algorithm}

\end{document}